\documentclass{article}

\PassOptionsToPackage{square, numbers}{natbib}

\usepackage[final]{neurips_2026}

\PassOptionsToPackage{square, numbers}{natbib}

\makeatletter
\renewcommand{\@noticestring}{Workshop on Who Verifies the Agents? Toward Reliable Agent Development, NeurIPS 2026.}
\makeatother

\usepackage[utf8]{inputenc} % allow utf-8 input
\usepackage[T1]{fontenc}    % use 8-bit T1 fonts
\usepackage{hyperref}       % hyperlinks
\usepackage{url}            % simple URL typesetting
\usepackage{booktabs}       % professional-quality tables
\usepackage{amsfonts}       % blackboard math symbols
\usepackage{nicefrac}       % compact symbols for 1/2, etc.
\usepackage{microtype}      % microtypography
\usepackage{xcolor}         % colors

\usepackage{graphicx}
\usepackage{multirow}
\usepackage{bold-extra}
\usepackage{amsmath}
\usepackage{yhmath}
\usepackage{placeins}

\title{Pdb or Not to Be: \\ Comparing Debugging Interfaces With Pass$_{X}$@$k$}
\workshoptitle{Who Verifies the Agents?}

\author{%
  Ryan Magnuson \\
  Code Metal \\
  \And
  Ellie Kitanidis \\
  Code Metal \\
  \And
  Ariel Lee \\
  Code Metal \\
  \AND \\
  \texttt{\{ryan.magnuson, ellie, ariel.lee\}@codemetal.ai} }

\begin{document}

\maketitle

\begin{abstract}
Interactive debuggers are indispensable to human software development, yet coding agents overwhelmingly rely on edit-execute-observe loops, using common bash tools, test scripts, and print statements to debug via repeated shell commands. Stepping through a debugger provides live access to runtime state, which may enable different problem-solving capabilities. However, prior work typically introduces the debugger as an additional tool or through a new subagent, making its individual contribution difficult to isolate from the associated scaffold changes. To address this, we compare a debugger-only agent against a bash-only agent on all 500 SWE-bench Verified instances, holding the rest of the scaffold fixed. We evaluate both architectures on two models, GPT-OSS-120B and Nemotron-3-Ultra-550B, with three independent runs per architecture $\times$ model configuration. Crucially, to distinguish interface specialization from expanded coverage under repeated sampling, we introduce pass$_X$@$k$, a cross-agent pass@$k$ that provides an unbiased estimator of the probability that at least one of $k$ agents is successful. Although the bash and debugger agents solve partially disjoint subsets of the benchmark in any single pair of runs, our pass$_X$@$2$ estimator reveals that this is largely due to run-to-run variance rather than tool complementarity. Across both models, we find that two bash rollouts outperform the combination of one bash and one debugger rollout. Our findings highlight the importance of accounting for stochastic variance in evaluating agent and tool diversity, and we hope that pass$_X$@$k$ will serve as a useful framework for the community.

\end{abstract}

\section{Introduction}
Most coding agents, including production-grade assistants \cite{anthropic_claudecode,openai_codex,github_copilot}, debug programs primarily through a shell: inserting print statements, executing code, and parsing outputs or exception stack traces. Although several works have demonstrated benefits from execution state inspection for agentic program repair \cite{zhong2024ldb, xiang2026adi, schuler2026tracegate}, live debuggers remain underexplored as an observation and execution-control interface, despite offering a direct path to runtime state information---live stack frames, bound values, and controlled stepping---that shell-only workflows typically must reconstruct through instrumentation and run artifacts. A central question is whether this interface enables qualitatively different debugging capabilities, or merely provides an alternative way to access information that a shell-only agent can already recover. Existing debugger-enabled agents typically attach the debugger to an existing scaffold by incorporating it as a first-class tool \citep{yuan2025debuggym, liu2025agenttd, liu2026swepdb} or as part of a multi-agent system \citep{wang2025inspectcoder, garg2026debug2fix, guo2026swedoctor}. Because the debugger is additive in these studies, it is difficult to disentangle the value of the debugger's runtime evidence from the complex impact that expanding the tool-call action space or multi-agent framework can have on the policy as a whole \citep{xu2026devil}. Comparative studies are further limited by the fact that agent performance is commonly reported for a single rollout \citep{zhang2024dei, wang2025inspectcoder, garg2026debug2fix}, despite the fact that membership in the solved task set can shift dramatically under repeated sampling of the same architecture \citep{mustahsan2025stochasticity, bjarnason2026randomness}. To distinguish potential specialization between tools from stochastic variance between runs, we define pass$_X$@$k$, a cross-agent equivalent of pass@$k$ that combines $n$ runs of each of $k$ agents into a single unbiased estimator of their combined solve rate. With this metric in hand, we ask: given a live debugger \emph{instead of} a shell, how is the agent's performance impacted, and is there evidence that the two interfaces provide complementary capabilities?

To answer this, we compare an agent equipped with an interactive Python debugger (\textbf{pdb-agent}) against one equipped with a general bash tool for debugging (\textbf{bash-agent}) on SWE-bench Verified \cite{jimenez2024swebench,openai2024swebenchverified}, with the model, prompts, budgets, sandbox, and all other tools held fixed. A single run of each agent misleadingly suggests that \textbf{pdb-agent} solves problems \textbf{bash-agent} misses. However, as Figure~\ref{fig:overlap} shows, the debugger-only agent's exclusive solves shrink in a repeat analysis with three runs per agent, indicating that much of its apparent specialization is sampling variance. Figure~\ref{fig:relay-curves} illustrates that two \textbf{bash-agent} rollouts outperform the combination of one \textbf{bash-agent} plus one \textbf{pdb-agent} rollout at every relay budget and under both models evaluated, while an ablation demonstrating that simultaneous access to both tools provides no apparent benefit to \textbf{bash-agent} is deferred to Appendix~\ref{app:pdb_shell}. Both the size of the performance gap and the token efficiency of the architecture are model-dependent, as shown in Figure~\ref{fig:token-budget}, with GPT-OSS paying roughly twice as much per solved problem under pdb. That said, the shell's generality cuts both ways; an interface that can execute anything is also difficult to contain, and in Section~\ref{sec:observations} we report an early run in which shell access exposed a benchmark leak. Taken together, our results provide a controlled, budget-matched comparison showing that, for these models, a live debugger neither matches a shell nor adds complementary coverage to one. We also contribute the pass$_X$@$k$ estimator, which lets any such cross-agent comparison be made with an unbiased, variance-aware statistic.

\section{Related work}

\textbf{Live debuggers for code repair agents.}
Prior work has demonstrated that exposing internal execution state to an agent, either through captured evidence \cite{zhong2024ldb, schuler2026tracegate} or agent-queryable interfaces \citep{xiang2026adi}, can improve its performance at program repair. Recently, several studies have explored integrating conventional suspended-process debuggers into agent scaffolds. Debug-gym \citep{yuan2025debuggym} compares an agent with ordinary code execution capabilities against the same agent with additional pdb capabilities, finding that even the strongest models struggle to fully leverage pdb commands. A concurrent work to ours, SWE-PDB~\cite{liu2026swepdb}, also reports low utilization of runtime inspection commands in base models with pdb access, while training significantly improves both effective tool use and performance. Debug2Fix \citep{garg2026debug2fix} retains a shell-capable repair agent and delegates pdb or JDB interaction to a specialized subagent equipped with a fixed set of debugger tools. Its ablations show that exposing the debugger tools directly to the main agent is ineffective, whereas delegation improves repair performance. SWE-Doctor \citep{guo2026swedoctor} uses a dedicated debugging agent to produce structured runtime diagnoses that guide a separate patch-generation agent. InspectCoder \citep{wang2025inspectcoder} similarly separates runtime diagnosis from patch generation, assigning pdb inspection and runtime perturbation to a specialized inspector agent. The authors report 33 benchmark instances solved by InspectCoder but by none of four baselines lacking live debugger access, though this comparison is based on a single trajectory per system, making it unclear if the exclusivity of these solves is robust to stochastic variation across independent runs.

\textbf{Tool and agent diversity.} Several recent works have proposed frameworks to take advantage of tool diversity through heterogeneous tool-use strategies \citep{chen2025tumix} or multi-tool aggregation \citep{yao2025multitag}. However, the effectiveness of an added tool depends not only on the new capabilities it provides, but also on how each tool is exposed and presented to the agent. \citep{xu2026devil} examines six different tool architectures with similar underlying capabilities and finds substantial differences in agent behavior, efficiency, and consistency between architectures. Our work provides an orthogonal analysis: we compare two tools with potentially different capabilities in an otherwise shared agent harness. Since each tool defines a distinct agent configuration in this setting, our analysis also relates to studies of agent diversity. \citep{zhang2024dei} compares the joint performance of $k$ different coding agents against $k$ runs of the same coding agent on SWE-bench Lite, and argues that inter-agent diversity leads to more uniquely solved instances than intra-agent sampling. However, the collective inter-agent performance of the $k$ agents relies on a single public leaderboard submission per agent rather than combining repeated independent runs per agent into an unbiased estimator.

\textbf{Using pass@\(k\) for agent trajectories.} Pass@\(k\) \cite{chen2021codex}, originally defined to evaluate single-shot code generation, estimates the probability that at least one of \(k\) sampled candidates is correct. In an agentic setting, each candidate is a complete interaction trajectory rather than a single completion. Repeated sampling can therefore be used to characterize performance in long-horizon software-engineering agents \cite{yang2024sweagent,pan2025swegym, bjarnason2026randomness, zhai2026doesrlexpandcapability}. \cite{zhai2026doesrlexpandcapability} furthermore introduces pass@$(k,T)$, which generalizes agentic pass@$k$ to condition also on interaction depth $T$. We define a similar depth-conditioned estimator in Section~\ref{sec:agent-evals} and extend the pass@$k$ formalism to describe the concept of a {cross-agent} pass$_X$@$k$.

\section{Methods and experiments}

\subsection{Experimental setting}
We evaluate on all 500 instances of SWE-bench Verified, the human-validated subset of SWE-bench \cite{jimenez2024swebench}. Each instance specifies a repository at a base commit, an issue description, and two test sets: \texttt{\textbf{FAIL\_TO\_PASS}} (F2P), which the repair must fix, and \texttt{\textbf{PASS\_TO\_PASS}} (P2P), which must not regress. Test files are restored to their benchmark state before grading to prevent solutions through test modification. We retain known problematic instances and count them as failures rather than filtering them \cite{openai2026swebench}.

We select GPT-OSS-120B \citep{openai2025gptoss} and Nemotron-3-Ultra-550B \citep{nvidia2026nemotron3ultra} as open-weight models we can deploy in secure environments. Each model is evaluated through vLLM using its default sampling configurations and medium reasoning effort. Each rollout executes in an isolated Podman container initialized from the same SWE-bench base repository state with git metadata sanitized to prevent access to the reference patch or commit history. Full provisioning, isolation, and infrastructure details are provided in Appendix~\ref{app:infra}.

\subsection{Agent architectures and rollout protocol}

\begin{figure}[h]
    \centering
    \makebox[\textwidth][c]{\includegraphics[width=1.05\textwidth]{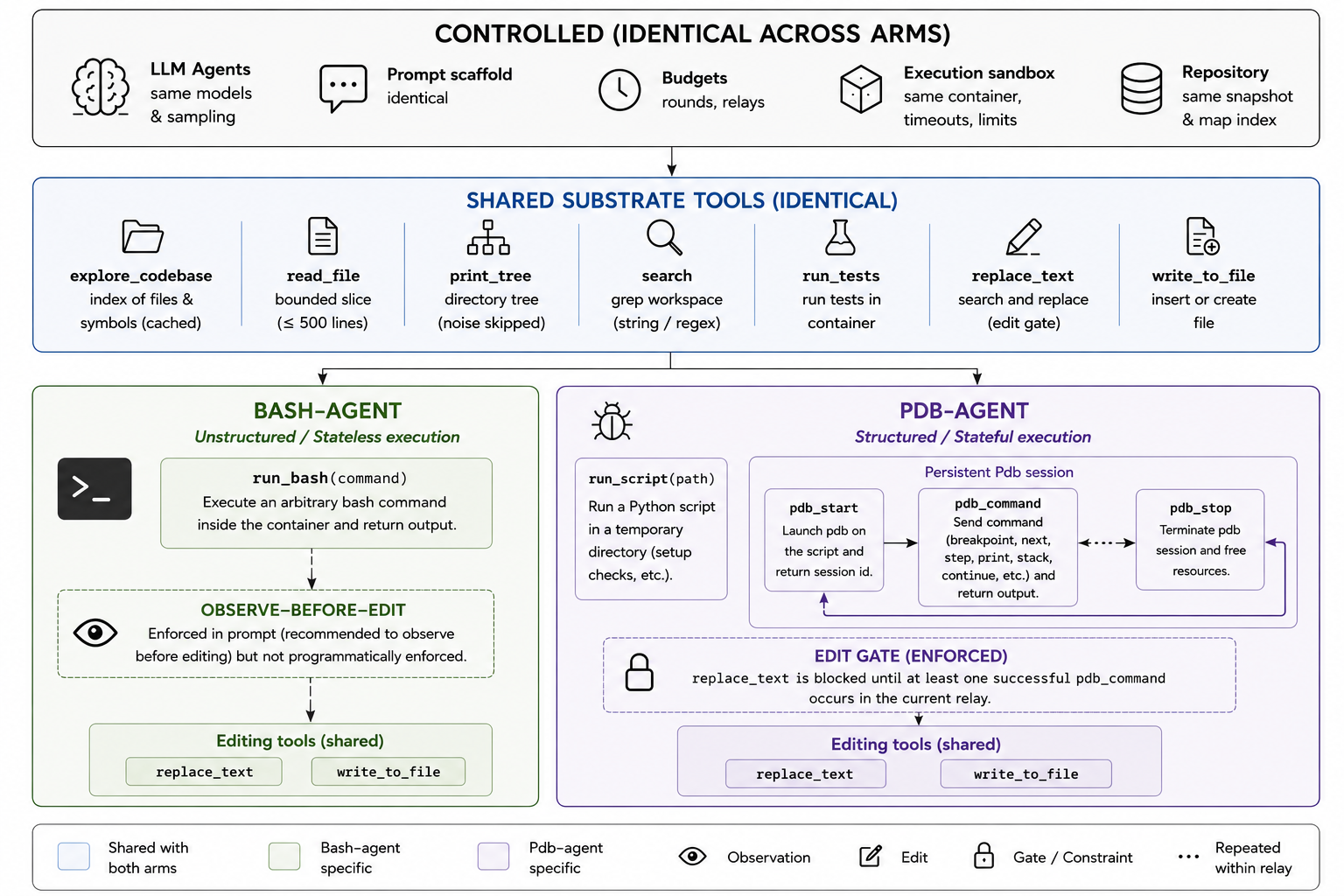}}
    \caption{\textbf{Controlled execution interface comparison.}  The two upper bands are identical across arms: models, sampling, prompts, relay and turn budgets, sandbox, repository snapshot, and a shared substrate of navigation, testing, and editing tools. Only the execution interface differs. The \textbf{bash-agent} runs arbitrary shell commands. The \mbox{\textbf{pdb-agent}} trades the shell for a debugger session plus a scoped script runner for setup, and its edit tools stay locked until one debugger command succeeds in the current relay. The gate is the one enforced constraint in the study, added after pilot runs showed the pdb arm editing blindly without ever inspecting state.}
    \label{fig:interfaces}
\end{figure}

We construct two matched agent architectures, \textbf{bash-agent} and \textbf{pdb-agent}, on a common harness, illustrated in Figure~\ref{fig:interfaces}. Across each architecture $\times$ model arm, we hold fixed the sampling configuration, budget, prompt scaffold, repository snapshot, sandboxes, a shared substrate of tools for navigation, testing, and editing, and a precomputed codebase map generated once per instance. 

The \textbf{bash-agent} executes arbitrary shell commands through \texttt{run\_bash}. The prompt requests observation before editing, but the harness does not enforce this constraint. The \textbf{pdb-agent} has no arbitrary shell access. It retains restricted Python execution for setup and interacts with a persistent debugger. Following the tool-limiting strategy of Debug2Fix \cite{garg2026debug2fix}, which disables file editing until the debugger has been called at least once, the \textbf{pdb-agent's} \texttt{replace\_text} is unavailable until at least one debugger command succeeds in the current relay. We adopted the gate after early pdb runs showed a recurring failure mode: the agent edited files without ever inspecting program state, then thrashed between the failing tests and successive blind edits until the relay budget was exhausted. Requiring one successful debugger observation before the first edit of each relay reduced this behavior in our pilot runs. We preferred the tool-level gate over stronger prompt instructions because it leaves the shared prompt scaffold lean and identical across arms. The \textbf{bash-agent} has no enforced equivalent beyond prompting. Each rollout consists of at most 10 relays, each containing at most 50 agent turns. At a relay boundary, a handoff step summarizes the current progress and initializes a fresh model context. As with standard compaction, workspace edits persist, but message history does not. We reinstate the edit gate at each relay in an effort to help prevent the flailing blind edits in the \textbf{pdb-agent}.

\subsection{Evaluating and comparing agents}
\label{sec:agent-evals}

For agent (architecture $\times$ model) $m$, task $j$, and run $i \in \{1,\dots,n\}$, we let

\begin{equation}
C_{ijm}
=
\mathbf{1}
\{
\text{run } i \text{ of agent } m \text{ is correct on task } j \}
\end{equation}

and define the usual pass@$k$ estimator, 

\begin{equation}
\widehat{\text{pass@}k} = 1 - \frac{\binom{n-c_{jm}}{k}}{\binom{n}{k}}, \ c_{jm} = \sum_{i=1}^n C_{ijm},
\end{equation}

which estimates the probability that at least one out of $k$ runs of agent $m$ will succeed on task $j$ (estimated by sampling $n \geq k$ runs of the same architecture). 

To evaluate the benefits of a diverse agent portfolio of $k$ agents against a baseline where the same compute is allocated to $k$ runs of a single agent, we define the \textbf{cross-pass-at-k} estimator, pass$_X$@$k$:

\begin{equation}
\widehat{\text{pass}_{X}\text{@}k} = 1 - \prod_{m=1}^k \frac{\binom{n-c_{jm}}{1}}{\binom{n}{1}} = 1 - \frac{1}{n^k}\prod_{m=1}^k ({n-c_{jm}}),
\label{eq:cross-pass-at-k}
\end{equation}

where the second term is the probability that all $k$ agents fail on task $j$. Appendix~\ref{app:unbiased_estimator} establishes unbiasedness under iid sampling within, and mutual independence between, each agent's set of runs.

In our work, we compare two agents and sample three runs per agent, so $n=3$\footnote{Our definition explicitly assumes each agent is sampled an equal number of times.} and $k=2$, and the intra- and inter-agent pass@2 are, respectively,

\begin{align}
    \widehat{\text{pass@2}}_{mm} &= 1 - \frac{\binom{3-c_{jm}}{2}}{3}  \label{eq:intra-pass2}\\
    \widehat{\text{pass}_{X}\text{@}2}_{mm^\prime} &= 1 - \frac{(3-c_{jm})(3-c_{jm^\prime})}{9},
    \label{eq:special-case}
\end{align}

where the latter captures the probability that a pair of (bash, pdb) runs will solve task $j$.

We can also threshold these statistics by the number of relays $r$, letting $C_{ijm} \rightarrow S_{ijm}(r)$:

\begin{align}
S_{ijm}(r) &= \mathbf{1} \{ \text{run } i \text{ of agent } m \text{ is correct on task } j \text{ and } R_{ijm} \leq r \}
\label{eq:eqn_6} \\
s_{jm}(r) &= \sum_{i=1}^n S_{ijm}(r)
\label{eq:eqn_7}
\end{align}

where $R_{ijm}$ is the number of relays used by this observed run \cite{zhai2026doesrlexpandcapability}.

\section{Results}

\subsection{Overall performance by model and architecture}
\label{sec:performance}

The final scores of each model $\times$ architecture arm is given in Table~\ref{tab:main} and plotted as a function of relay budget in Figure~\ref{fig:relay-curves}. With three independent runs for each of the four arms, we report pass@$k$ and its complement (pass\^{}$k$, the probability that all $k$ draws succeed) for $k=2$ and $k=3$. For pass@1, which is equivalent to the mean over the three runs, we also report the standard deviation.

\begin{table}[h!]
  \caption{Correctness rates by model and architecture, averaged over the entire dataset and with a maximum budget of $10$ relays. For pass@$1$ (i.e. the mean over three runs), we also report st. dev.}
  \label{tab:main}
  \centering
  \small
  \begin{tabular}{llccccc}
    \toprule
    \textbf{Model} & \textbf{Arch} & \textbf{pass@1}
      & \textbf{pass@2} &  \textbf{pass@3} & \textbf{pass\^{}2} & \textbf{pass\^{}3} \\
    \midrule
    \multirow{3}{*}{Nemotron-3-Ultra ($n{=}500$)}
        & bash & 56.7 $\pm$ 1.2 & 62.7 & 65.4 & 50.7 & 47.4 \\
        & pdb  & 49.6 $\pm$ 1.4 & 55.9 & 59.2 & 43.3 & 40.4 \\
        & cross & - & 60.3 & - & - & - \\
      \midrule
      \multirow{3}{*}{GPT-OSS-120B ($n{=}500$)}
        & bash & 56.7 $\pm$ 0.6 & 64.7 & 68.8 & 48.7 & 44.8 \\
        & pdb  & 40.3 $\pm$ 0.8 & 47.8 & 51.2 & 32.7 & 28.6 \\
        & cross & - & 60.0 & - & - & - \\
      \bottomrule
  \end{tabular}
\end{table}

Figure~\ref{fig:relay-curves} shows substantial performance gains from scaling sampling breadth $k$, an effect that remains consistent across observed relay depths and all architecture $\times$ model arms. Furthermore, \textbf{bash-agent} demonstrates a substantial lead over \textbf{pdb-agent} under both models, which perform comparably under this architecture. However, when only pdb tools are available, Nemotron-3-Ultra substantially outperforms GPT-OSS, and this margin is consistent across sampling breadth and relay depth. 

\begin{figure}[h!]
  \centering
  \includegraphics[width=0.6\linewidth]{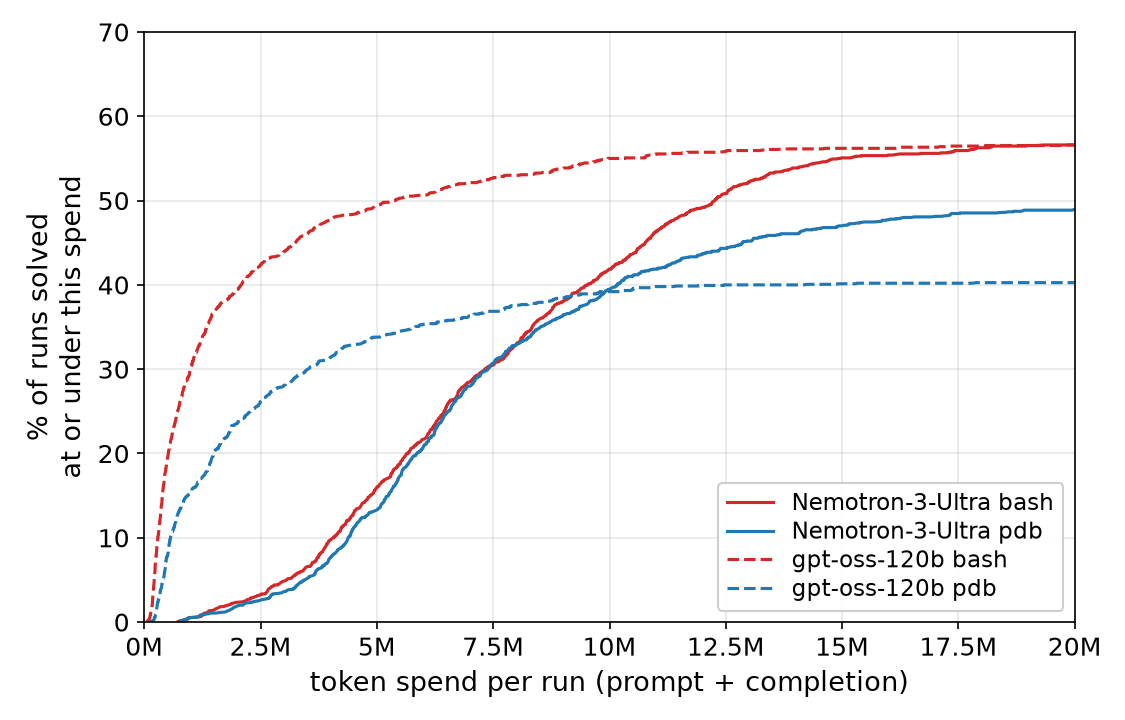}
  \caption{Share of runs solved at or under a given per-run spend threshold for each of the four arms.}
  \label{fig:token-budget}
\end{figure}

Analyzing performance as a function of token spend in Figure~\ref{fig:token-budget}, we see highly model-dependent efficiency curves, which may also inform the observed asymmetry in the pdb vs. bash gap between models. Nemotron's token spend is greater than GPT-OSS's (its median solved run costs several times more tokens) and scales similarly for either agent architecture, whereas the debugger agent roughly doubles the token cost per solved problem for GPT-OSS, which is able to exploit the bash tool relatively cheaply but quickly plateaus in performance as the token budget is increased.

\subsection{Cross-architectural performance by model}
\label{sec:cross-performance}

\begin{figure}[h!]
  \centering
  \includegraphics[width=0.75\textwidth]{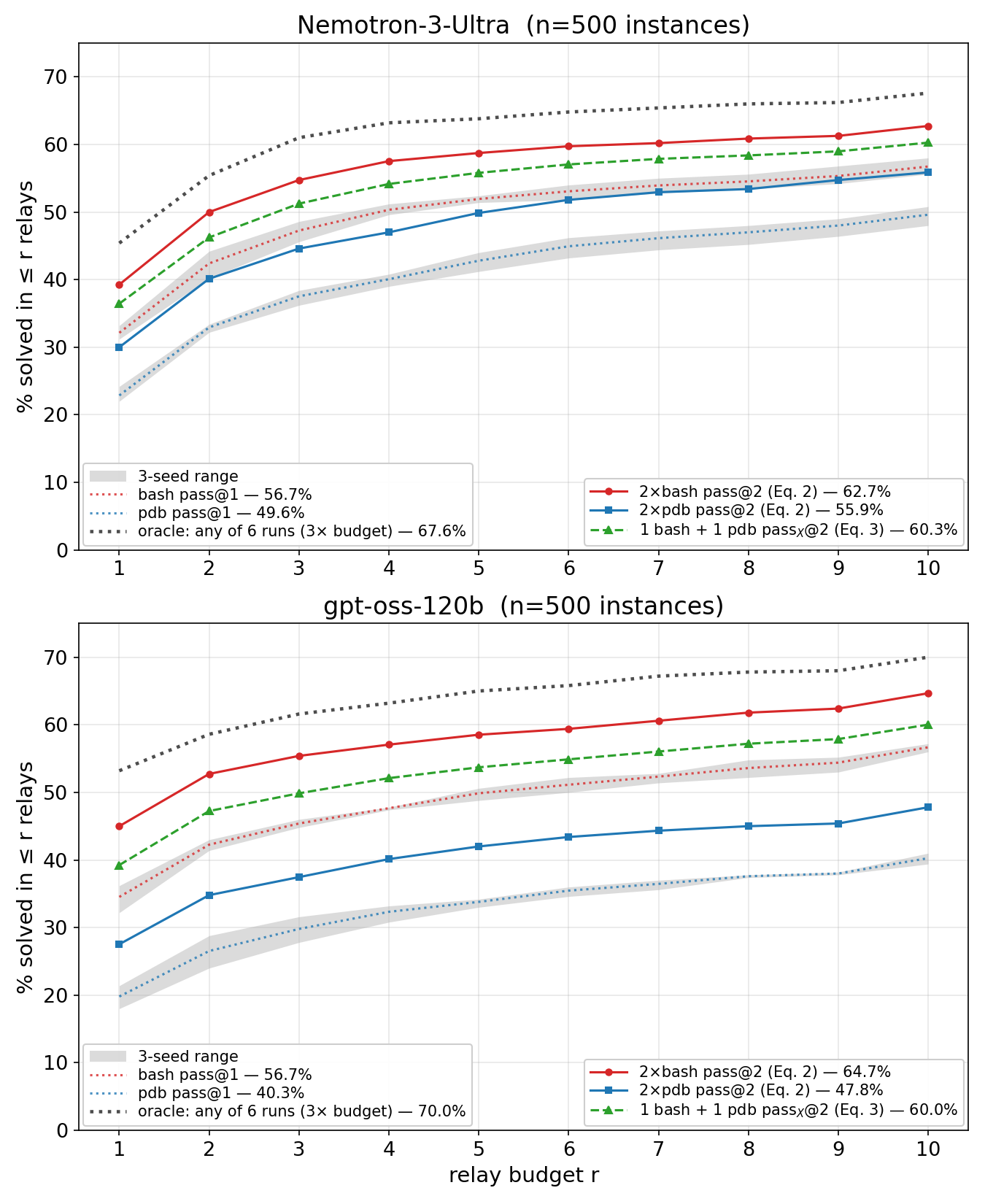}
\caption{Cumulative solve rate as a function of relay budget. Dotted curves are pass@1, with the shaded envelope spanning the three seeds. Solid curves are pass@2. Two bash rollouts (solid red) dominate the pass$_X$@2 of one bash + one pdb rollout (green, dashed) at every relay threshold. The oracle reranker (gray, dotted) unions all six runs at $3\times$ the budget and is plotted as a ceiling.}
  \label{fig:relay-curves}
\end{figure}

We also ask whether the two interfaces are complementary: for a given model, are there classes of problems where a debugger or a shell is the higher-leverage tool? In Figure~\ref{fig:relay-curves}, we see that the combination of a bash run and a pdb run (green triangles) consistently performs worse than two bash runs (red circles), a result that holds for both models and across all relay budgets.

\FloatBarrier

\begin{figure}[h!]
  \centering
  \includegraphics[width=\linewidth]{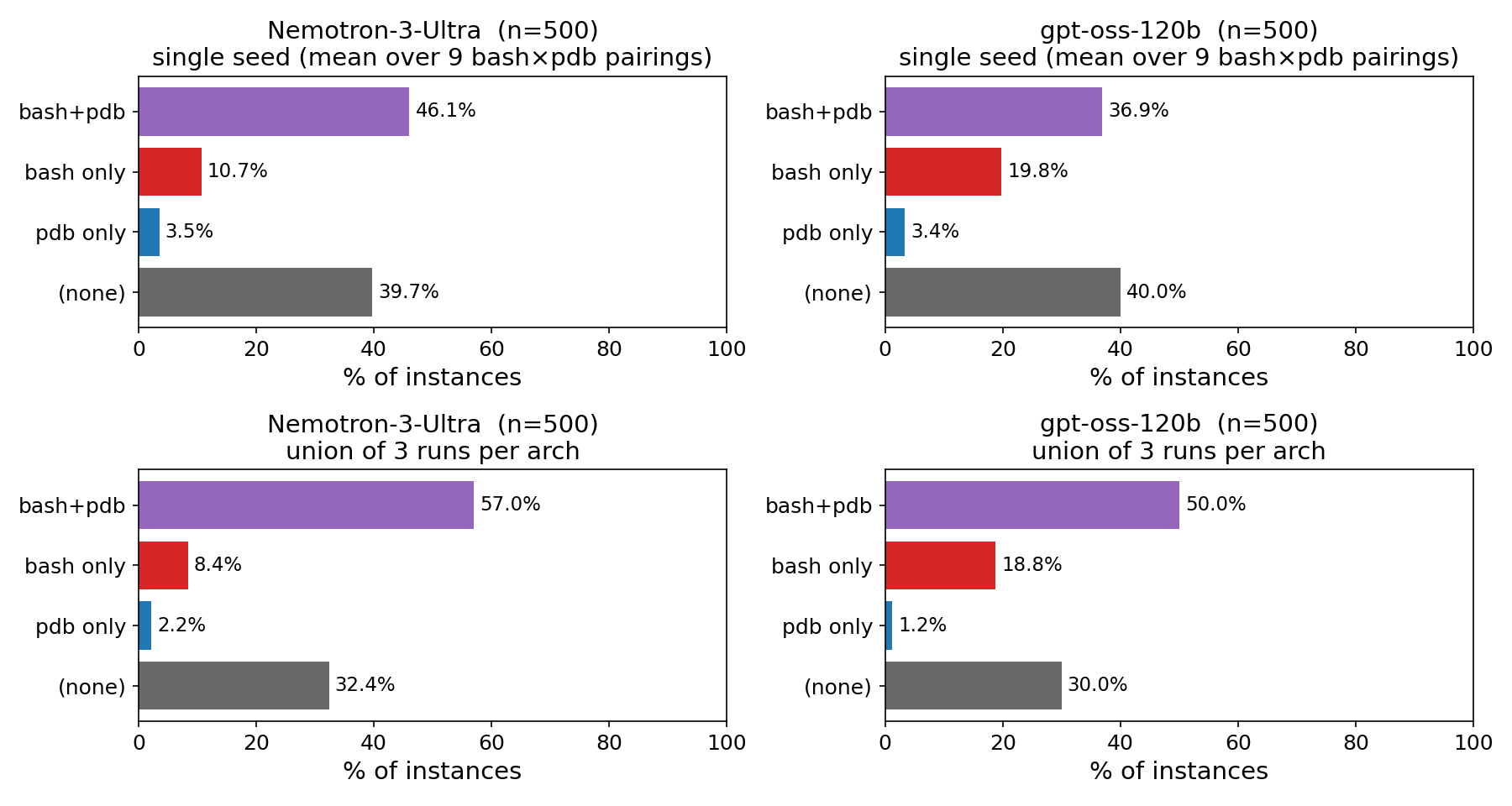}
  \caption{Who solved it. Top row: what a single-seed comparison sees (mean over all nine pairings). Bottom row: comparing the unions of the full three runs per architecture.}
  \label{fig:overlap}
\end{figure}

Figure~\ref{fig:overlap} highlights how the subset of instances that are seemingly solved by one agent but not the other shrinks substantially once all three seeds per arm are pooled. We conclude that any apparent specialization is mostly sampling variance between runs rather than a difference in what the two interfaces can solve. As an additional ablation, we test whether pass@1 is affected when adding the debugger alongside the shell, finding it fails to improve performance (see Appendix~\ref{app:pdb_shell}).

\subsection{Additional observations}
\label{sec:observations}

\paragraph{Complexity strata.} We stratify instances by gold-patch size (files touched and lines changed) as a rough proxy for bug complexity. As discussed in Appendix~\ref{app:strata}, pass rates fall under both arms as patches grow, but the bash--pdb gap shows no consistent trend. Most gold patches are small, with only 22 of the 500 instances in the 3+ file bucket and eight in the $>$100-line bucket, which limits our ability to conclusively determine whether debuggers help solve more complex bugs.

\paragraph{Shell access leak.} In an early run, the bash agent (Nemotron) used \texttt{git} to traverse the repository's commit history until it found the gold patch and then applied it. We subsequently sanitized every workspace (remotes removed, all refs but \texttt{HEAD} deleted, reflog expired) and added a tool-level guard on \texttt{git} invocations (see Appendix~\ref{app:infra}). The episode illustrates a structural property of the comparison: an unstructured shell exposes the entire environment, including artifacts the benchmark assumes hidden, so containment must be added piecemeal. 

\paragraph{Tool-name aliasing.} A recurring tool-use failure mode was the model calling a tool by a plausible name it does not have e.g. reaching for \texttt{search\_replace} when the registered tool is \texttt{replace\_text}. Registering aliases that map the names models actually reach for onto the intended tool converted these failed calls into successful ones at no cost to the interface, eliminating a class of wasted rounds. 

\section{Conclusions, limitations, and future work}
We investigate the effect of replacing, rather than adding, a bug-fixing agent's shell-based debugging channel with a live debugger, while holding all else equal. Despite the initial appearance of interface complementarity under a single-seed, Debug2Fix-style pilot study (described in more detail in Appendix~\ref{app:prelim}), we find that allocating a rollout to each architecture performs worse than allocating two rollouts to the \textbf{bash-agent} when evaluated under a budget-matched pass$_X$@$2$. However, our study is limited to two open-weight reasoning models and a single debugger; different models and debugging interfaces may yield different performance curves. An important confounding factor is the unknown extent to which each underlying model has been trained on particular tools, and on the benchmark itself: the Nemotron-3-Ultra technical report lists SWE-bench Verified among the public seed sets for its agentic post-training data, without specifying the split \cite{nvidia2026nemotron3ultra}, so some of its bash familiarity may be instance-specific rather than general. Additionally, SWE-bench Verified is a Python-only benchmark whose instances may under-represent the long-horizon, state-dependent failures that often necessitate execution-state inspection. SWE-bench Verified's complexity strata are too sparse at the hard end ($n{=}22$ for 3+ files, $n{=}8$ for $>$100 lines) to draw robust conclusions about relative performance on this sub-class of problems. In future work, we plan to expand our evaluation to more diverse tasks featuring stateful runtime failures, and to explore how our results change after fine-tuning the models on general pdb tool use.

\FloatBarrier

\bibliographystyle{plainnat}
\bibliography{references}

@misc{zhang2024dei,
  title        = {Diversity Empowers Intelligence: Integrating Expertise of
                  Software Engineering Agents},
  author       = {Zhang, Kexun and Yao, Weiran and Liu, Zuxin and Feng, Yihao and
                  Liu, Zhiwei and Murthy, Rithesh and Lan, Tian and Li, Lei and
                  Lou, Renze and Xu, Jiacheng and Pang, Bo and Zhou, Yingbo and
                  Heinecke, Shelby and Savarese, Silvio and Wang, Huan and
                  Xiong, Caiming},
  year         = {2024},
  eprint       = {2408.07060},
  archivePrefix= {arXiv},
  url          = {https://arxiv.org/abs/2408.07060}
}

@misc{xu2026devil,
  title        = {The Devil Is in the Interface: Evaluating How Tool Architecture
                  Shapes Coding Agent Behavior},
  author       = {Xu, Xiangzhe and Saghir, Hamidreza and Wu, Qianhui and
                  C{\^o}t{\'e}, Marc-Alexandre and Wang, Tong and
                  Lakkaraju, Kiran and Pei, Kexin and Zhang, Xiangyu},
  year         = {2026},
  eprint       = {2608.11386},
  archivePrefix= {arXiv},
  primaryClass = {cs.SE},
  url          = {https://arxiv.org/abs/2608.11386}
}

@misc{garg2026debug2fix,
  title        = {{Debug2Fix}: Can Interactive Debugging Help Coding Agents Fix
                  More Bugs?},
  author       = {Garg, Spandan and Huang, Yufan},
  year         = {2026},
  eprint       = {2602.18571},
  archivePrefix= {arXiv},
  primaryClass = {cs.SE},
  url          = {https://arxiv.org/abs/2602.18571}
}

@misc{yuan2025debuggym,
  title        = {debug-gym: A Text-Based Environment for Interactive Debugging},
  author       = {Yuan, Xingdi and Moss, Morgane and El Feghali, Charbel and
                  Singh, Chinmay and Moldavskaya, Darya and MacPhee, Drew and
                  Caccia, Lucas and Pereira, Matheus and Kim, Minseon and
                  Sordoni, Alessandro and C{\^o}t{\'e}, Marc-Alexandre},
  year         = {2025},
  eprint       = {2503.21557},
  archivePrefix= {arXiv},
  primaryClass = {cs.SE},
  url          = {https://arxiv.org/abs/2503.21557}
}

@article{xiang2026adi,
  title        = {Empowering Autonomous Debugging Agents with Efficient Dynamic Analysis},
  author       = {Xiang, Jiahong and Xu, Xiaoyang and Chu, Xiaopan and
                  Tian, Hongliang and Zhang, Yuqun},
  journal      = {Proceedings of the ACM on Software Engineering},
  volume       = {3},
  number       = {FSE},
  articleno    = {FSE031},
  year         = {2026},
  doi          = {10.1145/3797126},
  url          = {https://arxiv.org/abs/2604.24212}
}

@misc{zhong2024ldb,
  title        = {Debug like a Human: A Large Language Model Debugger via
                  Verifying Runtime Execution Step-by-step},
  author       = {Zhong, Li and Wang, Zilong and Shang, Jingbo},
  year         = {2024},
  eprint       = {2402.16906},
  archivePrefix= {arXiv},
  primaryClass = {cs.SE},
  url          = {https://arxiv.org/abs/2402.16906}
}

@misc{openai2026swebench,
  author       = {{OpenAI}},
  title        = {Why {SWE}-bench {V}erified No Longer Measures Frontier Coding Capabilities},
  howpublished = {OpenAI Research Blog},
  year         = {2026},
  month        = feb,
  note         = {Accessed: 2026-08-28},
  url          = {https://openai.com/index/why-we-no-longer-evaluate-swe-bench-verified/}
}

@inproceedings{
    jimenez2024swebench,
    title={{SWE}-bench: Can Language Models Resolve Real-world Github Issues?},
    author={Carlos E Jimenez and John Yang and Alexander Wettig and Shunyu Yao and Kexin Pei and Ofir Press and Karthik R Narasimhan},
    booktitle={The Twelfth International Conference on Learning Representations},
    year={2024},
    url={https://openreview.net/forum?id=VTF8yNQM66}
}

@inproceedings{yao2025multitag,
    title = "Diverse Multi-tool Aggregation with Large Language Models for Enhanced Math Reasoning",
    author = "Yao, Bohan  and
      Yadav, Vikas",
    editor = "Christodoulopoulos, Christos  and
      Chakraborty, Tanmoy  and
      Rose, Carolyn  and
      Peng, Violet",
    booktitle = "Findings of the {A}ssociation for {C}omputational {L}inguistics: {EMNLP} 2025",
    month = nov,
    year = "2025",
    address = "Suzhou, China",
    publisher = "Association for Computational Linguistics",
    url = "https://aclanthology.org/2025.findings-emnlp.1377/",
    doi = "10.18653/v1/2025.findings-emnlp.1377",
    pages = "25264--25282",
    ISBN = "979-8-89176-335-7"
}

@misc{yang2024sweagent,
  title        = {{SWE-agent}: Agent-Computer Interfaces Enable Automated Software
                  Engineering},
  author       = {Yang, John and Jimenez, Carlos E. and Wettig, Alexander and
                  Lieret, Kilian and Yao, Shunyu and Narasimhan, Karthik and
                  Press, Ofir},
  year         = {2024},
  eprint       = {2405.15793},
  archivePrefix= {arXiv},
  primaryClass = {cs.SE},
  note         = {NeurIPS 2024},
  url          = {https://arxiv.org/abs/2405.15793}
}

@article{liu2025agenttd,
  title={Agent That Debugs: Dynamic State-Guided Vulnerability Repair},
  author={Zhengyao Liu and Yunlong Ma and Jing-Xuan Xu and Junchen Ai and Xiang Gao and Hai-Long Sun and Abhik Roychoudhury},
  journal={ArXiv},
  year={2025},
  volume={abs/2504.07634},
  url={https://api.semanticscholar.org/CorpusID:277667352}
}

@article{wang2025inspectcoder,
  title={InspectCoder: Dynamic Analysis-Enabled Self Repair through interactive LLM-Debugger Collaboration},
  author={Yun-Kun Wang and Yue Zhang and Guo-Chang Li and Chen Zhi and Bin-Hua Li and Fei Huang and Yong-Bin Li and Shui-Guang Deng},
  journal={ArXiv},
  year={2025},
  volume={abs/2510.18327},
  url={https://api.semanticscholar.org/CorpusID:282246183}
}

@misc{mustahsan2025stochasticity,
      title={Stochasticity in Agentic Evaluations: Quantifying Inconsistency with Intraclass Correlation}, 
      author={Zairah Mustahsan and Abel Lim and Megna Anand and Saahil Jain and Bryan McCann},
      year={2025},
      eprint={2512.06710},
      archivePrefix={arXiv},
      primaryClass={cs.AI},
      url={https://arxiv.org/abs/2512.06710}, 
}

@article{bjarnason2026randomness,
  title={On Randomness in Agentic Evals},
  author={Bjarni Haukur Bjarnason and Andr{\'e} Silva and Martin Monperrus},
  journal={ArXiv},
  year={2026},
  volume={abs/2602.07150},
  url={https://api.semanticscholar.org/CorpusID:285453946}
}

@inproceedings{schuler2026tracegate,
author = {Schuler, Nicolas and Scotti, Vincenzo and Mirandola, Raffaela},
title = {TraceGate: A Framework for Policy-Controlled Failure Evidence in LLM-Assisted Program Repair},
year = {2026},
isbn = {9798400728822},
publisher = {Association for Computing Machinery},
address = {New York, NY, USA},
url = {https://doi.org/10.1145/3832783.3834596},
doi = {10.1145/3832783.3834596},
booktitle = {Proceedings of the 41st IEEE/ACM International Conference on Automated Software Engineering},
pages = {3369–3372},
numpages = {4},
location = {Munich, Germany},
series = {ASE '26}
}

@misc{guo2026swedoctor,
      title={SWE-Doctor: Guiding Software Engineering Agents with Runtime Diagnosis from Multi-Faceted Bug Reproduction Tests}, 
      author={Yaoqi Guo and Yang Liu and Jie M. Zhang and Yun Ma and Yiling Lou and Zhenpeng Chen},
      year={2026},
      eprint={2607.00990},
      archivePrefix={arXiv},
      primaryClass={cs.SE},
      url={https://arxiv.org/abs/2607.00990}, 
}

@misc{zhai2026doesrlexpandcapability,
      title={Does RL Expand the Capability Boundary of LLM Agents? A PASS@(k,T) Analysis}, 
      author={Zhiyuan Zhai and Wenjing Yan and Xiaodan Shao and Xin Wang},
      year={2026},
      eprint={2604.14877},
      archivePrefix={arXiv},
      primaryClass={cs.LG},
      url={https://arxiv.org/abs/2604.14877}, 
}

@article{chen2021codex,
  author  = {Mark Chen and Jerry Tworek and Heewoo Jun and Qiming Yuan and others},
  title   = {Evaluating Large Language Models Trained on Code},
  journal = {arXiv preprint arXiv:2107.03374},
  year    = {2021},
  url     = {https://arxiv.org/abs/2107.03374}
}

@inproceedings{pan2025swegym,
title = {Training Software Engineering Agents and Verifiers with SWE-Gym},
booktitle = {ICML},
author = {Jiayi Pan and Xingyao Wang and Graham Neubig and Navdeep Jaitly and Heng Ji and Alane Suhr and Yizhe Zhang},
year = {2025},
URL = {https://arxiv.org/abs/2412.21139}
}

@misc{chen2025tumix,
      title={TUMIX: Multi-Agent Test-Time Scaling with Tool-Use Mixture}, 
      author={Yongchao Chen and Jiefeng Chen and Rui Meng and Ji Yin and Na Li and Chuchu Fan and Chi Wang and Tomas Pfister and Jinsung Yoon},
      year={2025},
      eprint={2510.01279},
      archivePrefix={arXiv},
      primaryClass={cs.CL},
      url={https://arxiv.org/abs/2510.01279}, 
}

@misc{anthropic_claudecode,
  author       = {{Anthropic}},
  title        = {Claude Code},
  year         = {2024},
  note         = {Accessed: 2025-07-14},
  url          = {https://www.anthropic.com/claude-code}
}

@misc{github_copilot,
  author       = {{GitHub, Inc.}},
  title        = {{GitHub} {C}opilot: Your {AI} Pair Programmer},
  year         = {2024},
  note         = {Accessed: 2025-07-14},
  url          = {https://github.com/features/copilot}
}

@misc{openai_codex,
  author       = {{OpenAI}},
  title        = {{OpenAI} {C}odex},
  year         = {2025},
  note         = {Accessed: 2025-07-14},
  url          = {https://openai.com/codex/}
}

@article{nvidia2026nemotron3ultra,
  author       = {{NVIDIA}},
  title        = {Nemotron 3 {U}ltra: Open, Efficient Mixture-of-Experts
                  Hybrid {M}amba-Transformer Model for Agentic Reasoning},
  journal      = {arXiv preprint arXiv:2606.15007},
  year         = {2026},
  url          = {https://arxiv.org/abs/2606.15007}
}

@article{openai2025gptoss,
  author       = {{OpenAI}},
  title        = {gpt-oss-120b \& gpt-oss-20b Model Card},
  journal      = {arXiv preprint arXiv:2508.10925},
  year         = {2025},
  url          = {https://arxiv.org/abs/2508.10925}
}

@misc{openai2024swebenchverified,
    author       = {{OpenAI}},
    title        = {Introducing {SWE}-bench {V}erified},
    howpublished = {OpenAI Research Blog},
    year         = {2024},
    month        = aug,
    note         = {Accessed: 2026-08-29},
    url          = {https://openai.com/index/introducing-swe-bench-verified/}
}

@article{liu2026swepdb,
  author  = {Liu, Jiaxing and Hu, Xing and Xia, Xin},
  title   = {{SWE-PDB}: Teaching {LLMs} to Leverage Debugging Tools via Agentic Training},
  journal = {Proceedings of the ACM on Software Engineering},
  volume  = {3},
  number  = {ISSTA},
  year    = {2026},
  doi     = {10.1145/3832117}
}

\newpage

\appendix
\section{Complexity strata}
\label{app:strata}

  \begin{figure}[!htbp]
    \centering
    \includegraphics[width=0.9\linewidth]{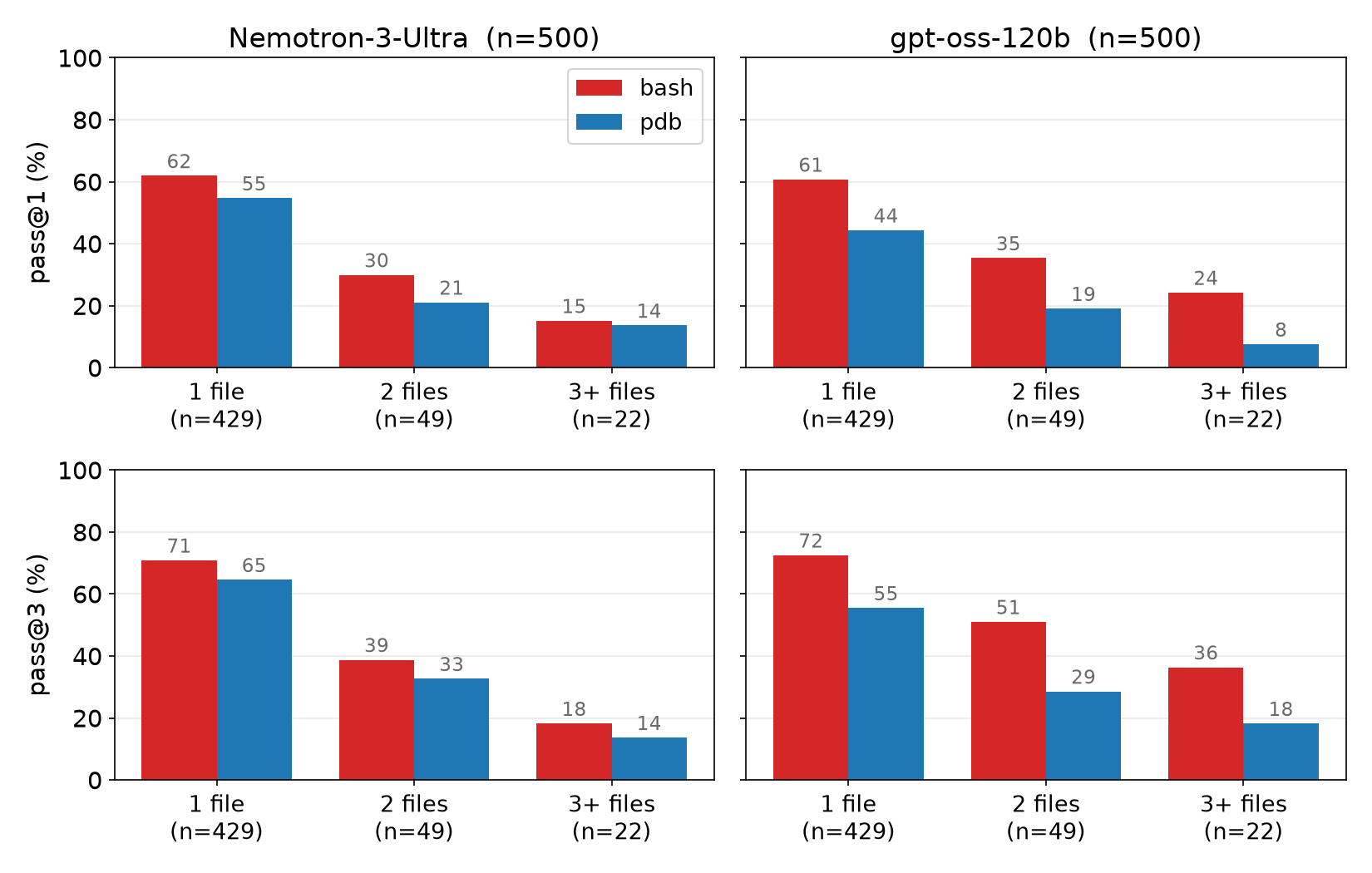}
    \caption{Pass rate by gold-patch complexity (files touched). Both interfaces decline in parallel; the debugger's deficit is roughly constant across buckets
  rather than shrinking on harder instances. The changed-lines stratification gives the same picture (Figure~\ref{fig:strata-lines}).}
    \label{fig:strata}

    \vspace{1em}

    \includegraphics[width=0.9\linewidth]{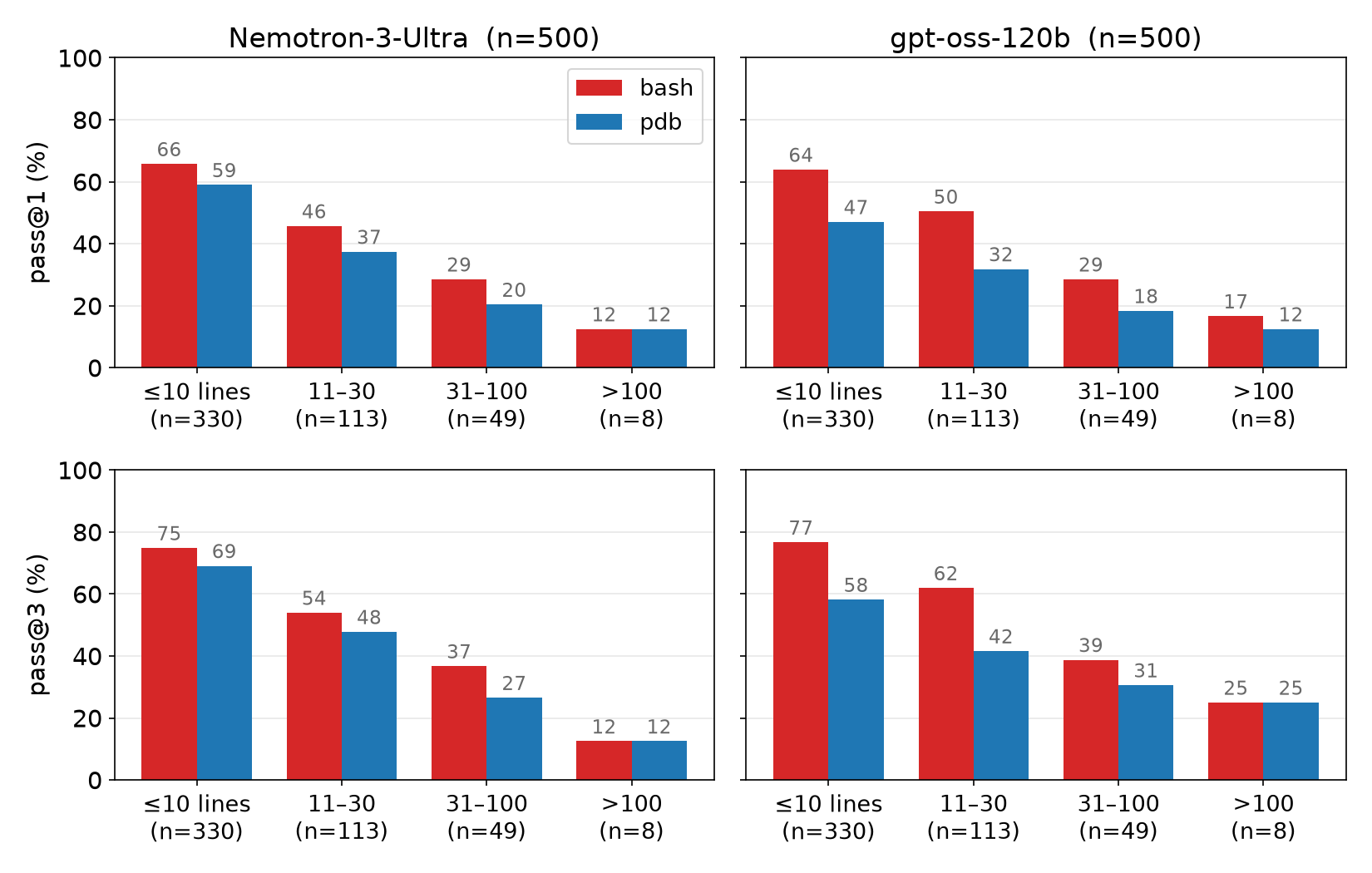}
    \caption{Pass rate by gold-patch complexity (changed lines). As with the files-touched stratification (Figure~\ref{fig:strata}), both interfaces decline in
  parallel as patches grow. The pdb--bash deficit is roughly stable through 100 lines and vanishes in the $>$100-line bucket under both models, but that bucket is
  too small to draw general conclusions, $n{=}8$.}
    \label{fig:strata-lines}
  \end{figure}

\section{Preliminary single-seed campaign}
\label{app:prelim}

\begin{figure}[h!]
  \centering
  \includegraphics[width=\linewidth]{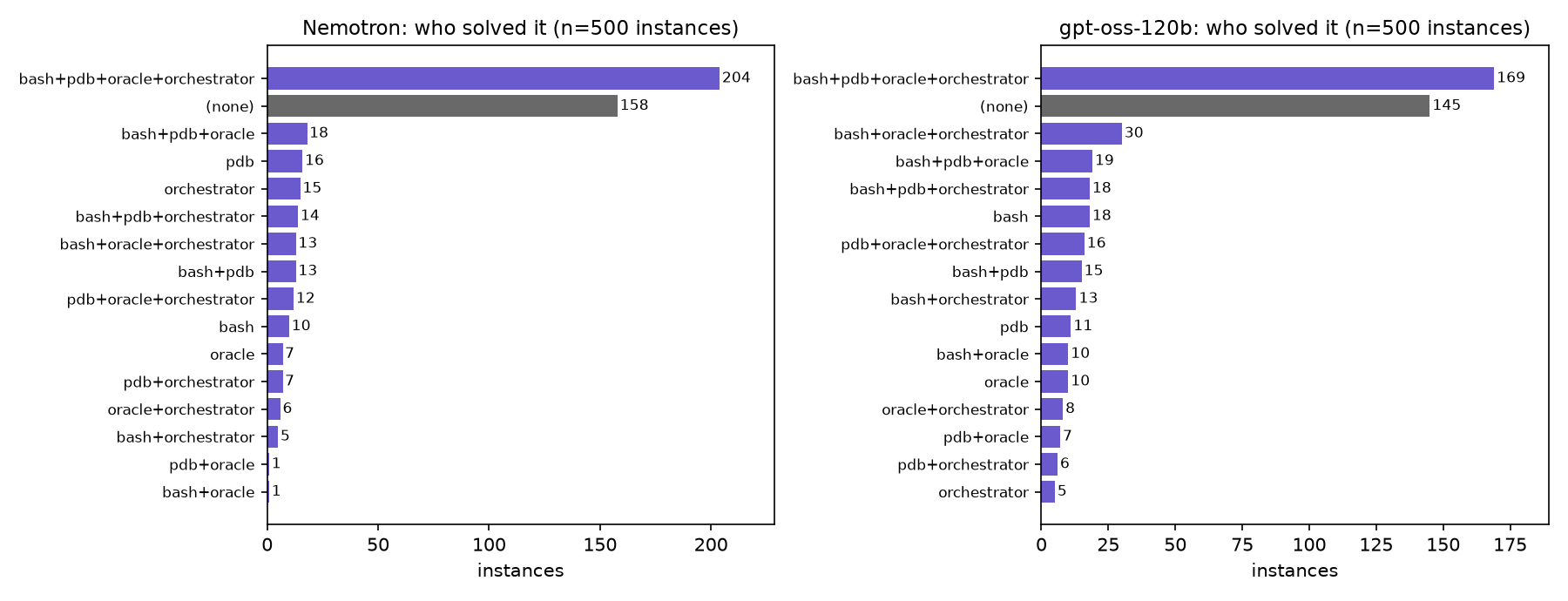}
  \caption{Which subset of the four architectures solved each instance in the preliminary single-seed campaign (Table~\ref{tab:prelim}). Both models are dominated by the two extremes---solved by all four arms (169 GPT-OSS, 204 Nemotron) or by none (145, 158)---while every exclusive pattern is small and no single architecture owns a large private set of solves. With one seed per arm, these small exclusive counts are consistent with the sampling variance quantified in Figure~\ref{fig:overlap} rather than with per-interface specialization.}
  \label{fig:who-solved-it}
\end{figure}

Before the main study, we ran a single-seed campaign over four architectures: \textbf{bash-agent}, \textbf{pdb-agent}, and two Debug2Fix-style \cite{garg2026debug2fix} compositions: an \textbf{oracle} agent that replaces direct debugger access with a dedicated debug subagent, and an \textbf{orchestrator} agent that additionally delegates file editing to an edit-specialized subagent. The results pointed in several directions at once (Table~\ref{tab:prelim}): the architecture ranking differed between models, with Nemotron's pdb arm outscoring its bash arm and the reverse for GPT-OSS, and every architecture owned a small set of exclusive solves (Figure~\ref{fig:who-solved-it}). Rather than ablate this design space until differences surfaced, we narrowed the study to the single controlled axis of the execution interface and measured it with seed-replicated runs. The seed-variance results of Section~\ref{sec:cross-performance} explain much of what this campaign appeared to show.

\begin{table}[h!]
    \caption{Resolve rates from the preliminary single-seed campaign: one run per (model, architecture, instance), four architectures. Rates are the fraction of instances resolved; with $n{=}1$ seed no variance estimate is available, and single-seed gaps of this size can flip between draws (\S\ref{sec:cross-performance}).}
    \label{tab:prelim}
    \centering
    \small
    \begin{tabular}{llcc}
      \toprule
      \textbf{Model} & \textbf{Arch} & \textbf{Solved} & \textbf{Rate (\%)} \\
      \midrule
      \multirow{4}{*}{Nemotron ($n{=}500$)}
        & bash & 278/500 & 55.6 \\
        & oracle & 262/500 & 52.4 \\
        & orchestrator & 276/500 & 55.2 \\
        & pdb & 285/500 & \textbf{57.0} \\
      \midrule
      \multirow{4}{*}{GPT-OSS-120B ($n{=}500$)}
        & bash & 292/500 & \textbf{58.4} \\
        & oracle & 269/500 & 53.8 \\
        & orchestrator & 265/500 & 53.0 \\
        & pdb & 261/500 & 52.2 \\
      \bottomrule
    \end{tabular}
\end{table}

\newpage
\section{Evaluation infrastructure and provisioning}
\label{app:infra}
We execute sweeps using a queue-driven EC2 fleet. SWE-bench instances are enqueued in Amazon SQS and consumed by workers launched from a common AMI containing the evaluation harness, Podman environment, and benchmark datasets. The scheduling unit is a (SWE-bench instance, model endpoint) pair: after claiming a message, a worker evaluates both agent architectures and all three repetitions for that instance against that endpoint, writes trajectory-level artifacts to S3, and claims the next available message. Workers terminate once the queue is exhausted, while interrupted jobs are returned to the queue for re-execution. Model inference is served independently through vLLM endpoints, making serving throughput rather than compute the primary concurrency constraint. We therefore choose parallelism empirically by increasing concurrent trajectories until aggregate token throughput saturates the available served models. Model inference was served on 12 NVIDIA B300s and 8 RTX 6000s over roughly 3 days of continuous operation, approximately 1{,}700 GPU-hours in total; cumulative agent-trajectory wall-clock across all runs was approximately 21{,}900 hours (912 days). All in, we processed 54.76 billion tokens through the runs---54.38B prompt + 386.3M completion.

Each instance is provisioned by cloning the repository, initializing at the base commit, applying the \texttt{test\_patch} from the dataset, and scoping the new tests. Applying the test patch at provisioning means the agent can read and run the failing tests during its attempt; this is deliberate, and departs from leaderboard-style evaluation. The per-run workspace's \texttt{git} history is sanitized after harness setup---the remote is removed, all refs except \texttt{HEAD} are deleted, and the reflog is expired---making the gold patch and all post-base-commit history unreachable from inside its run environment. A tool-level guard also exists triggering on any tool call containing git. Agents then operate on isolated per-run copies of the workspace in independent containers, run through Podman. Per-project test commands were incorporated into the \texttt{run\_tests} tool to ensure agents could run tests through the same interface regardless of the framework.

\section{Additive debugger arm: shell plus pdb}
\label{app:pdb_shell}

Debug2Fix~\cite{garg2026debug2fix}  reports that exposing debugger commands directly to a shell-capable agent is ineffective: in its single-run Java ablation, solve rate was flat for two of three models and dropped for the third, and the agent rarely invoked the tools. To check whether our substitution result is consistent with that additive finding, we ran a third architecture, \textbf{pdb\_shell}, that keeps the full \texttt{run\_bash} tool of the \textbf{bash-agent} and adds the persistent pdb session of the \textbf{pdb-agent}. The edit gate is not enforced in this arm. All other settings (models, sampling configuration, prompt scaffold, 10 relays $\times$ 50 turns, sandbox, codebase map) are identical to the two main arms, and we again perform three independently seeded runs over all 500 instances.

\begin{figure}[t]
  \centering
  \includegraphics[width=\linewidth]{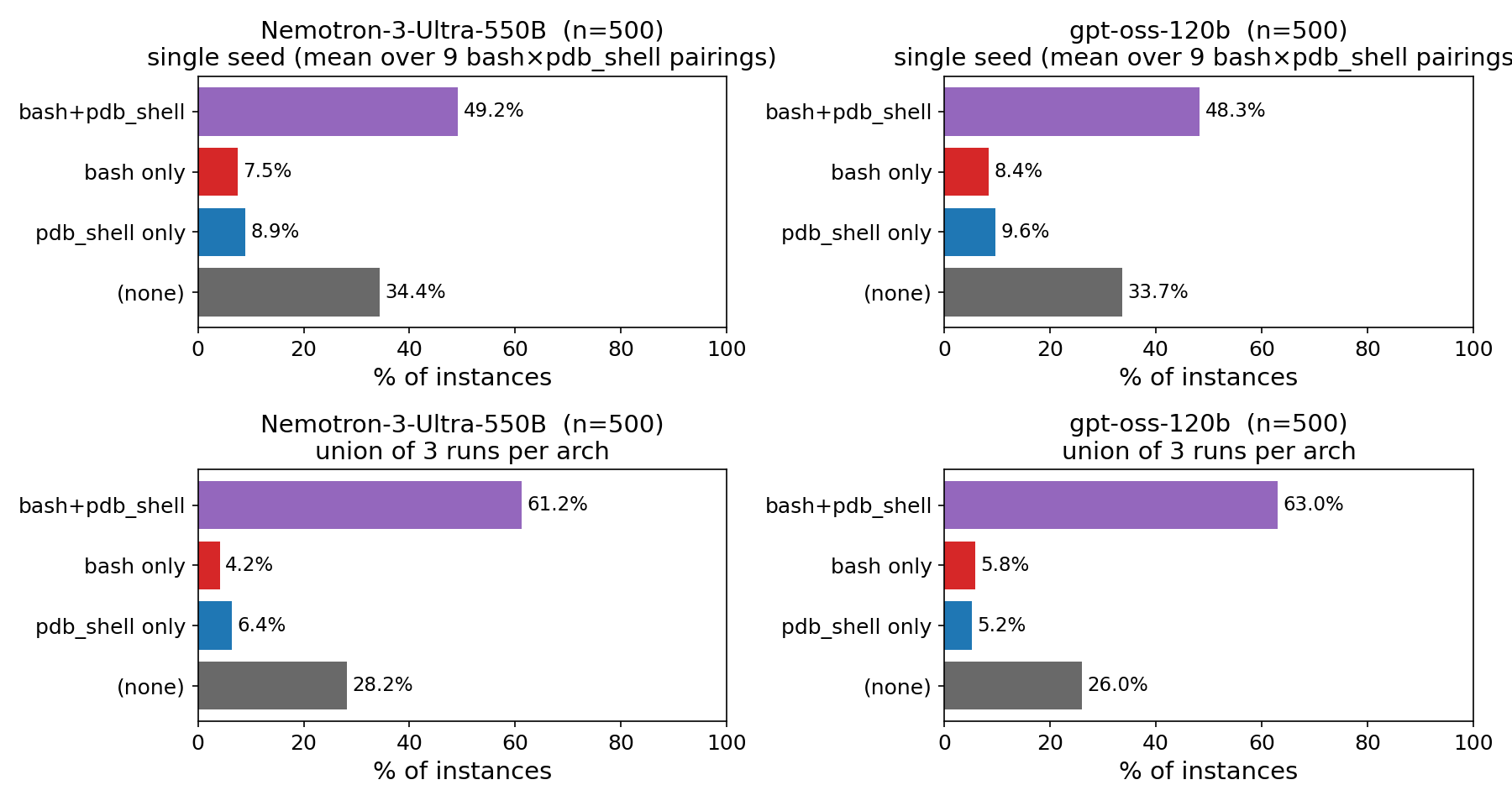}
  \caption{Who solved it, bash versus \textbf{pdb\_shell}, same construction as Figure~\ref{fig:overlap}. Top row: mean over the nine bash$\times$\textbf{pdb\_shell} seed pairings. Bottom row: union of all three runs per architecture. Unlike the substitution comparison, the exclusive-solve bars are symmetric between arms (union: 4.2\% bash-only vs.\ 6.4\% \textbf{pdb\_shell}-only for Nemotron; 5.8\% vs.\ 5.2\% for GPT-OSS-120B) and both shrink between rows, consistent with two arms of equal strength sampled independently.}
  \label{fig:overlap_pdb_shell}
\end{figure}

\begin{figure}[t]
  \centering
  \includegraphics[width=\linewidth]{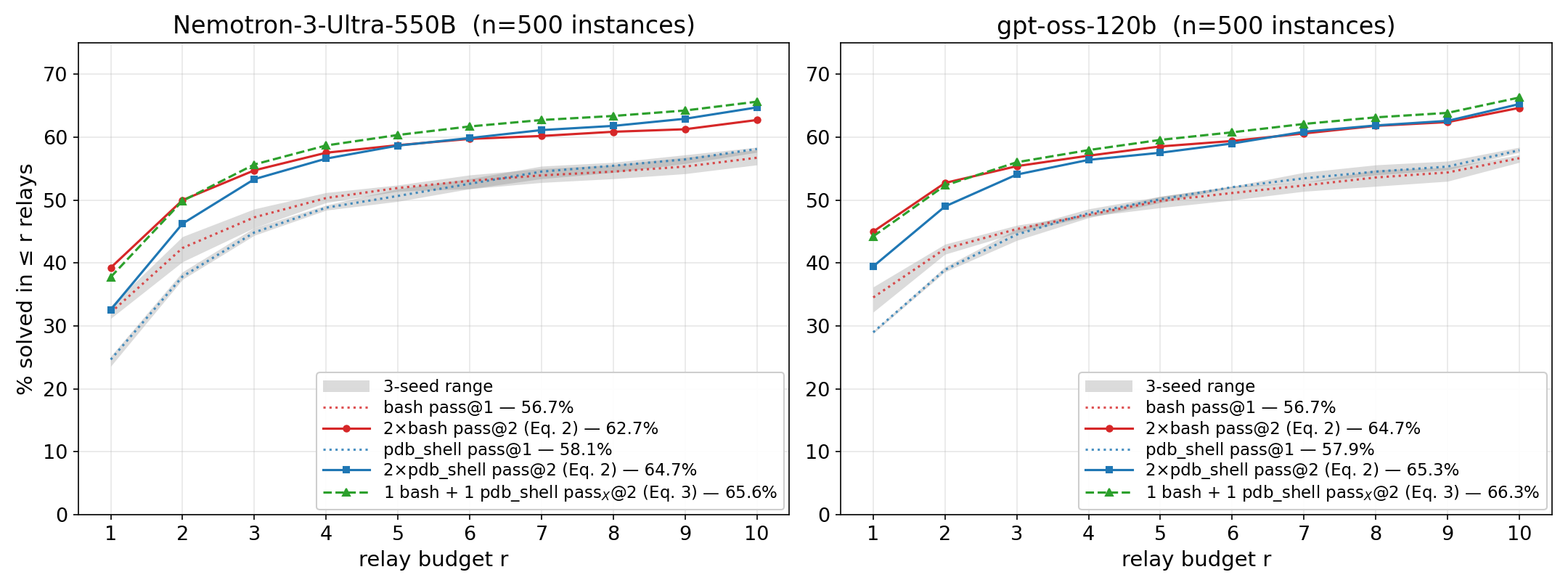}
  \caption{Cumulative solve rate as a function of relay budget $r$ for the additive \textbf{pdb\_shell} arm against \textbf{bash-agent}, same construction as Figure~\ref{fig:relay-curves}. Dotted curves are pass@1 with the 3-seed envelope; solid red and blue are intra-architecture pass@2 (Eq.~\eqref{eq:intra-pass2}); green is the cross-architecture pass@2 of one bash run and one \textbf{pdb\_shell} run (Eq.~\eqref{eq:special-case}). With the shell retained, \textbf{pdb\_shell} pass@1 matches bash pass@1 within about 1.5 points, and the mixed pair edges out both intra-architecture pairs at the full budget by roughly one point while trailing two bash runs at $r \leq 2$.}
  \label{fig:relays}
  \label{fig:relays_pdb_shell}
\end{figure}

Figure~\ref{fig:relays_pdb_shell} shows the result. Adding the debugger on top of the shell leaves pass@1 essentially unchanged: 58.1\% for Nemotron-3-Ultra and 57.9\% for GPT-OSS-120B against 56.7\% for \textbf{bash-agent} under both models, within about 1.5 points. This replicates the Debug2Fix observation that direct debugger access is roughly neutral, or destructive, for a shell-equipped agent, and it localizes the bash--pdb gap of Table~\ref{tab:main} to the removal of the shell rather than to the debugger itself. Under the budget-matched pass@2, the picture also differs from the substitution setting: the cross-architecture pairing of one bash run with one \textbf{pdb\_shell} run reaches 65.6\% (Nemotron) and 66.3\% (GPT-OSS-120B), marginally above two \textbf{pdb\_shell} runs (64.7\% and 65.3\%) and above two bash runs (62.7\% and 64.7\%). The margin over the stronger single arm is about one point under both models, and the mixed pair trails two bash runs at the smallest relay budgets ($r \leq 2$), so we read this as two near-equivalent arms whose exclusive solves are symmetric (Figure~\ref{fig:overlap_pdb_shell}) rather than as evidence of complementary abilities. When the debugger is additive, a diverse pair is at least as good an allocation as resampling either arm; when it replaces the shell, it is not.

\FloatBarrier

\section{Unbiasedness of cross-pass-at-k}
\label{app:unbiased_estimator}

We establish unbiasedness of the \textbf{cross-pass-at-k} estimator in Equation~\ref{eq:cross-pass-at-k} using the notation of Section~\ref{sec:agent-evals}. For a given task $j$ and fixed portfolio of $k$ agents, we assume that $C_{1jm},\dots,C_{njm}$ are independent and identically distributed (iid) for each agent $m$, and that the run batches are disjoint and mutually independent across agents.

% Fix a task $j$ and a prespecified portfolio of $k$ agents. The sample size $n \geq 1$ and evaluation configuration are fixed independently of the evaluation outcomes. Conditional on task $j$ and the fixed configuration, assume that $C_{1jm},...,C_{njm}$ are independent and identically distributed for each agent $m$, and that the run batches are disjoint and mutually independent across agents. 

\paragraph{Proposition.} Under these assumptions, $\widehat{\text{pass}_{X}\text{@}k}$ in Equation~\ref{eq:cross-pass-at-k} is an unbiased estimator of $ \Pr \left( \bigcup_{m=1}^k \{C_{ijm}=1\} \right)$, the probability that at least one of $k$ independently sampled runs, one per agent, succeeds.

\paragraph{Proof.} By the definition of $c_{jm}$ and identical distribution of runs within each agent's batch,
\begin{equation}
    \mathbb{E}[n-c_{jm}] = \sum_{i=1}^{n}\Pr(C_{ijm}=0)=n\Pr(C_{1jm} =0).
\end{equation}
Using Equation~\ref{eq:cross-pass-at-k},
\begin{equation}
    \begin{split}
        \mathbb{E}[\widehat{\text{pass}_{X}\text{@}k}] &= 1 - \frac{1}{n^k}\mathbb{E}\left[\prod_{m=1}^k ({n-c_{jm}})\right] \\
        &= 1 - \frac{1}{n^k}\prod_{m=1}^k\mathbb{E}[n-c_{jm}] \\
        &= 1 - \prod_{m=1}^k\Pr(C_{ijm}=0) \\
        &= \Pr \left( \bigcup_{m=1}^k \{C_{ijm}=1\} \right).
    \end{split}
\end{equation}

Thus, 
\begin{align}
        \text{Bias}(\widehat{\text{pass}_{X}\text{@}k}) = \mathbb{E}[\widehat{\text{pass}_{X}\text{@}k}] - \Pr \left( \bigcup_{m=1}^k \{C_{ijm}=1\} \right) = 0.
\end{align}

Averaging over a fixed benchmark preserves unbiasedness by linearity of expectation, and independence across tasks is not required. For any fixed relay threshold $r$, the same argument applies upon replacing $C_{ijm}$ by $S_{ijm}(r)$ and $c_{jm}$ by $s_{jm}(r)$ from Equations~\eqref{eq:eqn_6}--\eqref{eq:eqn_7}, provided the sampling assumptions hold for the complete run records. This yields pointwise unbiasedness for the probability that at least one run succeeds within the per-run relay threshold.

\end{document}